\documentclass[letterpaper]{article} 
\usepackage{aaai2027}  
\usepackage{amsmath}
\usepackage[hyphens]{url}  
\usepackage{graphicx} 
\usepackage{natbib}  
\usepackage{caption} 
\usepackage{multirow}
\usepackage{algorithm}
\usepackage{algorithmic}
\usepackage{newfloat}
\usepackage{placeins}
\usepackage{listings}
\DeclareCaptionStyle{ruled}{labelfont=normalfont,labelsep=colon,strut=off} 
\floatstyle{ruled}
\newfloat{listing}{tb}{lst}{}
\floatname{listing}{Listing}

\usepackage{booktabs}

\title{EchoChange: A Diffusion Language Model with Dual Pass Remasking for Factual Remote Sensing Disaster Change Captioning}
\author{
    Dongwei Sun\textsuperscript{\rm 1}\equalcontrib,
    Bowen Yao\textsuperscript{\rm 2}\equalcontrib,
    Yujie Zhang\textsuperscript{\rm 2},
    Pei Liu\textsuperscript{\rm 1},
    Jing Yao\textsuperscript{\rm 3},
    Xiangyong Cao\textsuperscript{\rm 1}
}
\affiliations{
    \textsuperscript{\rm 1}School of Computer Science and Technology and the Ministry of Education Key Lab for Intelligent Networks and Network Security, Xi'an Jiaotong University, 710049, China\\
    \textsuperscript{\rm 2}School of Computer Science and Technology, Faculty
of Electronic and Information Engineering, Xi’an Jiaotong University, Xi’an
710049, China\\
    \textsuperscript{\rm 3}State Key Laboratory of Remote Sensing and Digital Earth, Aerospace Information Research Institute, Chinese Academy of Sciences, Beijing, 100094, China\\

    \{sundongwei, BowenYao2007, zhang\_yujie2005\}@outlook.com, lpei53682@gmail.com, jasonyao92@gmail.com, caoxiangyong@xjtu.edu.cn 
}

\begin{document}

\maketitle

\begin{abstract}

Bi-temporal remote-sensing disaster change captioning often needs to identify sparse and spatially localized changes across large pre- and post-event scenes and then translate them into coherent, factual descriptions. However, existing change captioning methods always follow an autoregressive decoding paradigm to generate the change description and thus an early misinterpretation of the changed object, event, or spatial relation becomes an irreversible premise for subsequent text, amplifying visual ambiguity into cascading factual errors. To address this limitation, we propose EchoChange, a multimodal discrete diffusion language model that formulates change captioning as iterative masked-token denoising rather than left-to-right generation. By repeatedly revising the entire caption while conditioning on the image pair, EchoChange can reconsider uncertain content and correct imperfect intermediate predictions. We further introduce draft-aware dual-pass training, a progressive masking curriculum, and confidence-guided remasking to align training with iterative inference. Extensive experiments on the RSCC benchmark show that EchoChange substantially outperforms both general-purpose and remote-sensing-specific baselines across lexical and semantic metrics. The EchoChange Project is at \url{https://sundongwei.github.io/EchoChange_Project/}

\end{abstract}


\section{Introduction}

Rapid post-disaster assessment requires semantic interpretation beyond merely locating altered pixels. Analysts must identify what changed, where it occurred, and how the post-event scene differs from its pre-event state. Bi-temporal remote-sensing (RS) change captioning supports this goal by translating image pairs into descriptions of objects, events, quantities, and spatial relations. Early work established the task, while subsequent Transformer-based methods and larger benchmarks broadened its scope \cite{Hoxha2022ChangeCaptioning,Liu2022RSICCformer,sun2026scnet}. Disaster assessment, however, imposes a stricter requirement: captions must be factually accurate, not merely fluent.

\begin{figure}[t]
  \centering
  \includegraphics[width=\linewidth]{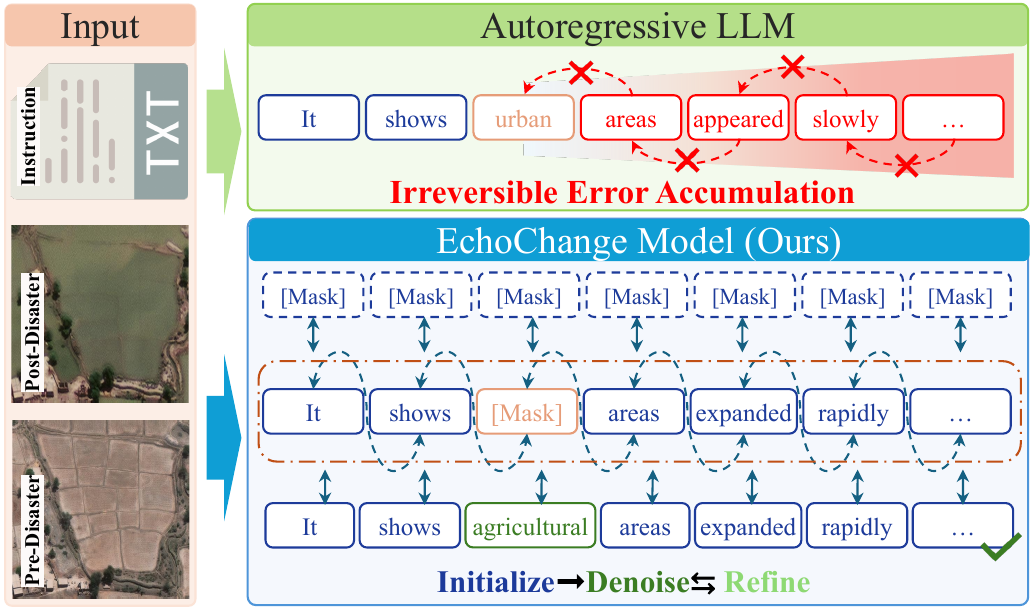}
  \caption{Generation under identical multimodal conditions. Autoregressive decoding commits to each generated token, allowing early factual errors to influence later predictions. EchoChange instead remasks uncertain tokens and iteratively refines an editable answer.}
  \label{fig:demo}
\end{figure}

Factual captioning remains difficult because event evidence is sparse, while illumination, seasonal variation, sensor differences, and misregistration can mimic genuine change. General-purpose multimodal large language models (MLLMs) and RS assistants improve temporal visual-language reasoning, but most still decode captions autoregressively \cite{Wang2024CCExpert,Irvin2025TEOChat}. An early factual error therefore becomes part of the prefix that conditions every later prediction, making the output difficult to revise. Teacher-forced training compounds this problem by exposing the decoder to gold prefixes during training but its own predictions during inference \cite{bengio2015scheduled}. Figure~\ref{fig:demo} illustrates how this mismatch can propagate a local comparison error into a globally inconsistent description.

Iterative masked prediction and diffusion language models offer a revisable alternative, allowing uncertain tokens to be regenerated from global context \cite{ghazvininejad2019maskpredict,austin2021d3pm,li2022diffusionlm,gong2023diffuseq,sahoo2024mdlm}. Existing RS diffusion methods do not fully realize this possibility. Diffusion-RSCC denoises continuous word embeddings, whereas Mask Approximation Net diffuses a visual change mask but retains autoregressive caption decoding \cite{Yu2025DiffusionRSCC,Sun2025MaskApproximationNet}. This leaves a central question: \textit{can multimodal caption generation be formulated as discrete token-space denoising that explicitly revises uncertain factual hypotheses?}

We answer this question with \textbf{EchoChange}, a multimodal discrete diffusion language model for bi-temporal disaster change captioning. EchoChange represents the caption as an editable answer region and constructs it through masked-token denoising conditioned on the ordered pre- and post-event images. At each iteration, the model jointly predicts all editable positions, retains high-confidence content, and remasks uncertain positions for reconsideration against the visual pair and the evolving caption. Intermediate predictions of disaster entities, events, quantities, and relations can thus be revised after global caption context becomes available.

Masked denoising alone, however, does not teach a model to correct its own output. Conventional single-pass training corrupts a reference caption while leaving every visible token copied from the ground truth. Iterative inference instead conditions on the model's preceding draft, whose visible tokens may already contain factual or structural errors. EchoChange bridges this train--inference gap through dual-pass remasking. Pass~1 predicts the entire answer region from a corrupted reference and forms a model-generated draft. After low-confidence positions are remasked, Pass~2 reconstructs the reference from this draft without restoring the retained positions from the ground truth. The second pass must therefore recover missing content, revise incorrect visible tokens, and preserve content already supported by the images. Curriculum timestep sampling expands the training distribution from local token recovery to near-fully masked caption generation, while confidence calibration makes the uncertainty scores used for remasking more reliable. During inference, a two-stage schedule first establishes the global change structure through denoising and then applies the learned draft-repair behavior to the nearly complete caption during polishing. These components align training with iterative inference and allow EchoChange to generate and revise intermediate hypotheses within a unified process.

We evaluate EchoChange on the RSCC disaster benchmark \cite{Chen2025RSCC} against eleven general-purpose and remote-sensing-specific baselines. EchoChange ranks first across the reported lexical-overlap and semantic-similarity metrics. Controlled correction experiments further show that it can revise single and multiple factual errors across diverse change categories, while the clean-control evaluation measures its ability to preserve valid descriptions. Fragment-recovery experiments and inference-stage ablations separately examine masked-content reconstruction and the complementary roles of denoising and polishing. 

Our main contributions are threefold:
\begin{itemize}
    \item We formulate bi-temporal RS disaster change captioning as multimodal discrete masked-token diffusion, giving the generated caption a globally editable representation rather than fixing tokens in left-to-right order.
    \item We introduce dual-pass remasking to bridge the clean-context gap between diffusion training and iterative inference, supported by curriculum timestep sampling, confidence calibration, and confidence-guided two-stage refinement.
    \item We compare EchoChange with eleven general-purpose and remote-sensing-specific baselines and evaluate generation, factual correction, clean-caption preservation, and masked-fragment recovery, demonstrating strong caption quality and measurable draft-revision capability.
\end{itemize}

\section{Related Work}

\paragraph{RS Change Captioning and Multimodal Disaster Understanding.}
Change captioning was first studied as describing differences between paired natural images, with later work emphasizing robustness to viewpoint distractors and localized visual changes \cite{jhamtani2018differences,park2019robust}. In remote sensing, foundational studies established dedicated datasets and cross-temporal architectures \cite{Hoxha2022ChangeCaptioning,sun2024light}; subsequent methods improved attentive change localization, task decoupling, spatial--temporal modeling, and semantic guidance \cite{chang2023chg2cap,liu2023promptcc,liu2024rscama,zhu2024semanticcc}. Domain-aligned vision--language pretraining and MLLMs further broadened RS interpretation \cite{liu2024remoteclip,kuckreja2024geochat,muhtar2025lhrsbot,zhang2024earthgpt,Wang2024CCExpert,Irvin2025TEOChat}, while disaster benchmarks emphasized building damage, post-flood understanding, and detailed temporal descriptions \cite{gupta2019xbd,rahnemoonfar2021floodnet,Chen2025RSCC}. Nevertheless, studies of image captioning and MLLMs show that fluent outputs may remain visually ungrounded \cite{rohrbach2018object,dai2023plausible,li2023pope}. Decoding-time methods mitigate such hallucinations through visual contrast or retrospection \cite{leng2024vcd,huang2024opera}, but retain autoregressive text generation. Diffusion-RSCC instead denoises continuous caption embeddings, whereas Mask Approx Net diffuses a visual change mask while keeping an autoregressive caption decoder \cite{Yu2025DiffusionRSCC,Sun2025MaskApproximationNet}. EchoChange differs by making discrete caption tokens the diffusion state, so the textual hypothesis itself remains editable.

\paragraph{Non-Autoregressive and Diffusion Text Generation.}
Non-autoregressive generation removes strict left-to-right dependence \cite{gu2018nonautoregressive}, and iterative-refinement models showed that complete sequences can be repeatedly revised \cite{lee2018iterative,ghazvininejad2019maskpredict}. Diffusion then provided probabilistic formulations over categorical and general discrete state spaces \cite{hoogeboom2021multinomial,austin2021d3pm}. Text-oriented models have explored continuous embeddings \cite{li2022diffusionlm,gong2023diffuseq}, vocabulary simplices \cite{han2023ssdlm,karimimahabadi2024tess}, absorbing or soft-mask corruption \cite{he2023diffusionbert,chen2023maskeddiffuse,zhou2024diffusionnat,sahoo2024mdlm}, and score-entropy objectives \cite{lou2024sedd}. Asada and Miwa further train discrete diffusion models on self-generated intermediate states to reduce the training--inference discrepancy \cite{asada2025traininference}. EchoChange extends this revisable-generation line to bi-temporal multimodal conditioning: intermediate tokens are assessed against paired RS observations, low-confidence regions are selectively remasked, and its evaluation directly measures correction of controlled factual errors.

\section{Methodology}

\subsection{Problem Formulation and Overview}

Given a pre-disaster image $I^{1}$, a post-disaster image $I^{2}$, and an instruction $q$, EchoChange generates a factual change description $\mathbf{y}$. The description must identify disaster-related entities and actions while preserving quantities, severity, and spatial relations. Unlike autoregressive models, which commit to each token once it is emitted, EchoChange maintains an editable answer region and treats generation as conditional discrete denoising. Before denoising, we use CLIP to calculate an answer-length estimate from the image pair:
\begin{equation}
\begin{aligned}
\hat{L}&=\operatorname{CLIPLen}(I^{1},I^{2}),\\
\hat{\mathbf{y}}&\sim
p_{\theta}(\mathbf{y}_{1:\hat{L}}\mid I^{1},I^{2},q).
\end{aligned}
\label{eq:problem}
\end{equation}
where $\hat L$ is stored as \texttt{pred\_len} and determines the number of answer slots at inference. $\operatorname{CLIPLen}$ denotes a direct CLIP-based calculation, it introduces neither trainable parameters nor an auxiliary length loss. Training uses the ground-truth answer length, whereas inference obtains no length information from the reference caption.

\begin{figure*}[t]
    \centering
\includegraphics[width=\linewidth]{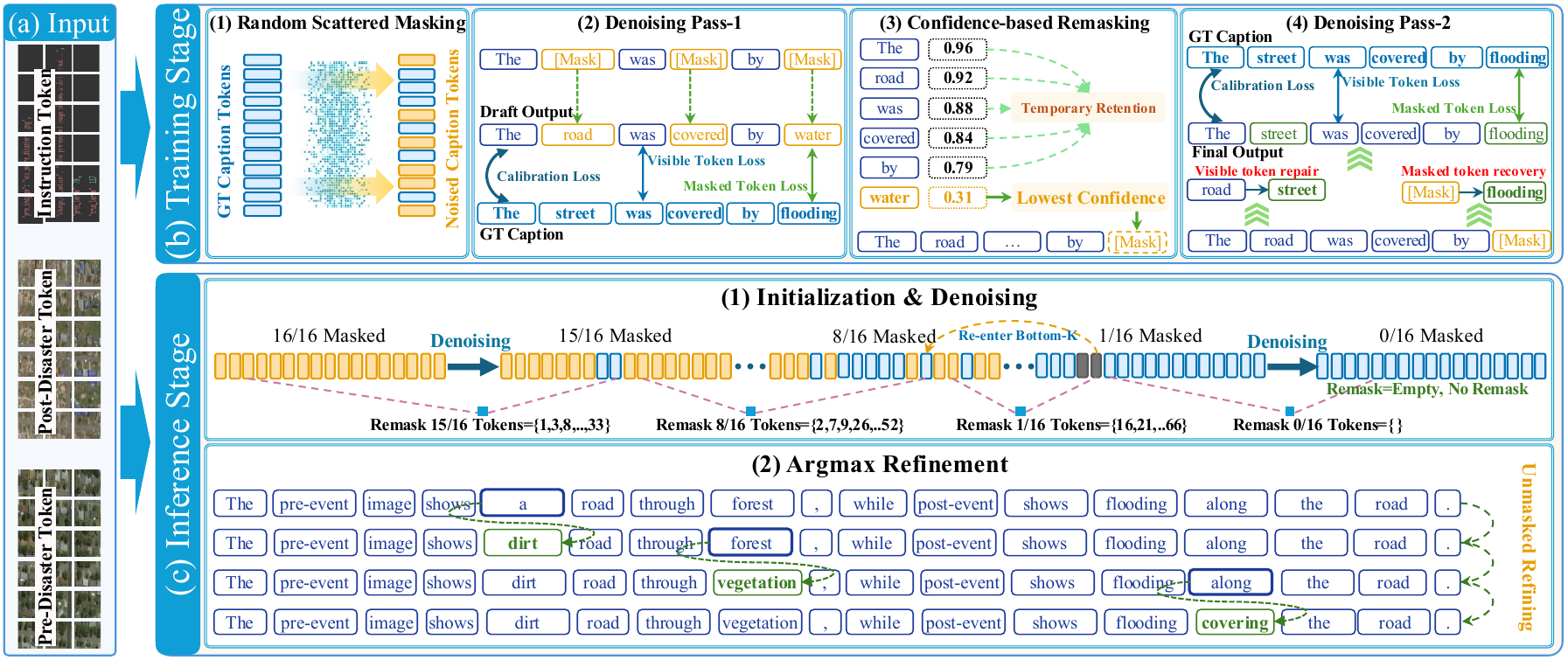}
    \caption{Overview of EchoChange. (a) The pre- and post-disaster images and task instruction remain fixed conditions. (b) During training, only ground-truth caption tokens are corrupted with 
[Mask] to form a curriculum-controlled mask set. Pass 1 produces a self-generated draft; its least-confident masked positions are then remasked, and both passes are supervised using the original masked and visible target sets. (c) Standard inference starts from a fully masked answer slot, retains confident predictions through scheduled denoising, and then applies mask-free polishing. The dashed branch denotes imperfect-caption initialization used only for correction analysis.}
    \label{fig:echochange_overview}
\end{figure*}

EchoChange contains three connected parts as shown in Figure ~\ref{fig:echochange_overview}. The multimodal backbone first encodes the ordered pre- and post-disaster images as fixed visual conditions. The model then corrupts only the assistant answer and learns to reconstruct it under a curriculum that increases the masking difficulty. A second pass replaces the clean ground-truth context with the model's own draft, so even visible answer tokens can be erroneous and must be checked against the images and global sentence context. At inference, the CLIP-derived length initializes a fully masked answer region, after which confidence-guided denoising and a short polishing stage repeatedly apply this learned draft-repair behavior. The standard generation, correction, clean-control, and fragment-recovery experiments use the same editable answer interface, which allows generation, revision, preservation, and reconstruction to be tested separately.
\subsection{Bi-Temporal Visual Conditioning}

EchoChange uses the native multimodal encoder to process the two remote-sensing images. The input template identifies $I^{1}$ as the pre-disaster observation and $I^{2}$ as the post-disaster observation, preserving their temporal order when the processor converts them into visual tokens. We denote the resulting representations by $\mathbf{V}^{1}$ and $\mathbf{V}^{2}$ and write the complete conditional sequence as
\begin{equation}
\mathbf{C}=[\mathbf{V}^{1},\mathbf{V}^{2},\mathbf{q}],
\qquad
\mathbf{x}=[\mathbf{C},\mathbf{y}].
\label{eq:multimodal_condition}
\end{equation}
The prompt and both sets of visual tokens remain visible during every training and inference pass; only positions in $\mathbf{y}$ can be masked or rewritten. EchoChange relies on the backbone's native cross-image reasoning and applies its task-specific changes to the language-generation process.

\subsection{Answer-Region Discrete Diffusion Learning}

\paragraph{Answer-only corruption.}
The multimodal prompt, image tokens, and instruction remain visible throughout training. EchoChange applies noise only to the assistant answer, preventing the model from treating missing visual evidence as part of the generation task. Let $a_i$ mark answer positions and let $\rho_t$ denote the masking ratio at diffusion step $t$. For each answer token, we sample $b_i^{(t)}\sim\mathrm{Bernoulli}(\rho_t)$. The masked set $M_t$, visible-answer set $V_t$, and corrupted input share one definition:
\begin{equation}
\label{eq:answer_corruption}
\begin{aligned}
M_t
    &= \bigl\{i \mid a_i = 1,\ b_i^{(t)} = 1\bigr\}, \\
V_t
    &= \bigl\{i \mid a_i = 1,\ b_i^{(t)} = 0\bigr\}, \\
\widetilde{x}_i^{(t)}
    &=
    \begin{cases}
        \texttt{[Mask]}, & i \in M_t, \\
        x_i,              & i \notin M_t.
    \end{cases}
\end{aligned}
\end{equation}
We use pure mask corruption rather than random token replacement. The training input therefore matches the inference trajectory, which starts from a masked answer region.

\paragraph{Curriculum timestep sampling.}
The curriculum expands the accessible noise range as training proceeds. Let $\tau\in[0,1]$ denote normalized progress within the curriculum ramp and $u\sim\mathcal{U}(0,1)$. We sample the timestep and mask ratio by
\begin{equation}
\begin{aligned}
t_{\max}(\tau)
    &= \min\!\left\{
        T,\,
        t_{\max}^{(0)}
        + \tau\bigl(T - t_{\max}^{(0)}\bigr)
    \right\}, \\
t
    &= t_{\min}
        + \operatorname{round}\!\left(
        u^{\gamma_{\tau}}
        \bigl[t_{\max}(\tau) - t_{\min}\bigr]
    \right), \\
\rho_t
    &= \frac{t}{T}.
\end{aligned}
\label{eq:curriculum}
\end{equation}
The exponent $\gamma_{\tau}$ decreases from $1.0$ to $0.5$, biasing later sampling toward larger timesteps. After the ramp, the scheduler sets $t=T$ with probability $p_{\mathrm{full}}=0.1$, exposing the model to the near-fully-masked state used at inference. The schedule connects local token recovery with caption-level generation without introducing a different objective for each noise level.

\paragraph{Dual-pass Remasking.}
A single denoising pass is trained in a clean-context regime: every unmasked answer token is copied from the ground truth. Iterative inference is different because its visible context comes from the preceding model prediction and may contain an incorrect entity, event, count, relation, or sentence structure. A model trained only in the former regime can become a strong masked-token infiller without learning to revise its own mistakes. EchoChange narrows this train--inference gap with two forward passes. Pass~1 predicts the entire answer region and forms a self-generated draft $\bar{\mathbf{y}}$. The model then remasks the $k_t$ least-confident positions from the initial masked set $M_t$ and sends the resulting draft to Pass~2:
\begin{equation}
\label{eq:dual_pass}
\begin{aligned}
\mathbf{P}^{(1)}
    &= f_{\theta}\!\left(
        \mathbf{C}, \widetilde{\mathbf{y}}^{(t)}
    \right), \\
c_i
    &= \max_{w} P_i^{(1)}(w), \\
\bar{y}_i
    &= \arg\max_{w} P_i^{(1)}(w),
    \qquad i \in M_t \cup V_t, \\
k_t
    &= \left\lceil
        \rho_{t'}\,\lvert M_t\rvert
    \right\rceil, \\
R_t
    &= \operatorname{BottomK}\!\left(
        M_t, \{c_i\}_{i\in M_t}, k_t
    \right), \\
\mathbf{P}^{(2)}
    &= f_{\theta}\!\left(
        \mathbf{C},
        \operatorname{Remask}\!\left(
            \bar{\mathbf{y}}, R_t
        \right)
    \right).
\end{aligned}
\end{equation}
Here $t'<t$ denotes the next target noise level. The remasking budget $k_t$ therefore gives Pass~2 more editable positions at high noise and fewer near the end of the schedule. Importantly, Pass~2 receives model predictions at every answer position rather than restoring $V_t$ from the ground truth. Hence, a retained token can satisfy $\bar y_i\neq y_i$ even when $i\notin R_t$. Both passes nevertheless use the original sets $M_t$ and $V_t$ for supervision; $R_t$ changes the second input but does not redefine its targets. Copying an incorrect draft token therefore incurs loss, forcing the model to validate visible content against the bi-temporal evidence and bidirectional sentence context. This is the key difference between masked-token completion and draft-level correction.

\paragraph{Compact training objective.}
The objective combines token reconstruction with confidence calibration. For pass $s\in\{1,2\}$ and answer subset $A\in\{M_t,V_t\}$, we define
\begin{equation}
\begin{aligned}
\widehat{y}_i^{(s)}
    &= \arg\max_{w} P_i^{(s)}(w), \\
c_i^{(s)}
    &= \max_{w} P_i^{(s)}(w), \\
\ell_{i,s}^{\mathrm{cal}}
    &=
    \begin{cases}
        \bigl(c_i^{(s)}\bigr)^2,
            & \widehat{y}_i^{(s)} \neq y_i, \\
        \bigl(1-c_i^{(s)}\bigr)^2,
            & \widehat{y}_i^{(s)} = y_i,
    \end{cases} \\
\mathcal{L}_{\mathrm{CE},s}^{A}
    &= -\frac{1}{\lvert A\rvert}
       \sum_{i\in A}
       \log P_i^{(s)}(y_i), \\
\mathcal{L}_{\mathrm{cal},s}^{A}
    &= \frac{1}{\lvert A\rvert}
       \sum_{i\in A}
       \ell_{i,s}^{\mathrm{cal}}.
\end{aligned}
\end{equation}
Masked-token cross-entropy drives factual recovery. The lower-weight visible-token term has a dual role: it teaches Pass~2 to replace erroneous retained draft tokens, while discouraging unnecessary changes when the visible context is already correct. Calibration penalizes confident errors as well as under-confident correct predictions. Both passes are combined as
\begin{equation}
\label{eq:training_objective}
\begin{aligned}
\mathcal{L}
= \sum_{s=1}^{2} w_s \Bigl[
    &\lambda_m\,\mathcal{L}_{\mathrm{CE},s}^{M_t}
     + \lambda_v\,\mathcal{L}_{\mathrm{CE},s}^{V_t} \\
    &+ \lambda_c\Bigl(
        \mathcal{L}_{\mathrm{cal},s}^{M_t}
        + \lambda_v\,\mathcal{L}_{\mathrm{cal},s}^{V_t}
      \Bigr)
\Bigr].
\end{aligned}
\end{equation}
The main configuration sets $w_1=w_2=1$ and $(\lambda_m,\lambda_v,\lambda_c)=(1.0,0.2,0.1)$, assigning lower weight to visible-token supervision. Calibration is necessary because the inference procedure uses confidence to decide which factual tokens can remain and which must be revised. 

\subsection{Length-Aware Confidence-Guided Inference}

EchoChange first calculates $\hat L$ from the pre- and post-disaster images using CLIP. This deterministic step requires no learned length head and produces the \texttt{pred\_len} used by the generator. EchoChange then creates $\hat L$ \texttt{[Mask]} tokens as the answer region while keeping the visual evidence and instruction fixed. The model receives no reference-derived length at inference and must fill the CLIP-sized answer region from the image pair.

The denoising stage updates all editable positions in parallel. Let $E$ contain the editable answer positions and let $r_k$ denote the remasking ratio at iteration $k$. Starting from a fully masked answer, one iteration is
\begin{equation}
\label{eq:iterative_inference}
\begin{aligned}
\widehat{\mathbf{y}}^{(K)}
    &= \operatorname{Repeat}\!\left(
        \texttt{[Mask]}, \widehat{L}
    \right), \\
\mathbf{P}^{(k)}
    &= f_{\theta}\!\left(
        \mathbf{C}, \widehat{\mathbf{y}}^{(k)}
    \right), \\
c_i^{(k)}
    &= \max_{w} P_i^{(k)}(w), \\
r_k
    &= \frac{t_{k-1}}{T}, \\
R_k
    &= \operatorname{BottomK}\!\left(
        E,\,
        \{c_i^{(k)}\}_{i\in E},\,
        \left\lceil r_k\,\lvert E\rvert\right\rceil
    \right), \\
\widehat{y}_i^{(k-1)}
    &=
    \begin{cases}
        \texttt{[Mask]},
            & i \in R_k, \\
        \displaystyle\arg\max_{w} P_i^{(k)}(w),
            & i \in E\setminus R_k.
    \end{cases}
\end{aligned}
\end{equation}
The ratio $r_k$ decreases with the timestep. Early iterations can revise broad sentence structure, while later iterations focus on uncertain entities, change actions, counts, or relations. A token can therefore be replaced after the model has observed the rest of the draft, rather than remaining fixed because it appeared early in the sequence.

Once the remasking schedule reaches zero, a short polishing stage performs full-answer refinement without introducing new masks. This stage is not a separately trained post-processor. Its capability follows from Pass~2 training, where the model learns to predict the ground truth from its own complete or partially remasked draft and cannot assume that a visible token is correct. Denoising first establishes the global factual structure; polishing then reuses this learned draft editor to remove residual factual or lexical inconsistencies and incomplete phrases. 

\begin{algorithm}[ht]
\caption{EchoChange training and inference.}
\label{alg:echochange}
\begin{algorithmic}[1]
\REQUIRE Image $(I^{1},I^{2})$; instruction $q$;
Ref. $\mathbf{y}$ for training
\ENSURE Predicted change description $\hat{\mathbf{y}}$

\STATE $(\mathbf{V}^{1},\mathbf{V}^{2})
\leftarrow \operatorname{EncodePair}(I^{1},I^{2})$
\STATE $\mathbf{C}\leftarrow
[\mathbf{V}^{1},\mathbf{V}^{2},\mathbf{q}]$

\STATE \textbf{Training}
\STATE $t\leftarrow\operatorname{CurriculumSample}()$
\STATE $(\widetilde{\mathbf{y}}^{(t)},M_t,V_t)
\leftarrow\operatorname{MaskAnswer}(\mathbf{y},t)$
\STATE $\mathbf{P}^{(1)}
\leftarrow f_{\theta}
(\mathbf{C},\widetilde{\mathbf{y}}^{(t)})$
\STATE $\bar{\mathbf{y}}\leftarrow
\operatorname*{arg\,max}\mathbf{P}^{(1)}$
\STATE $R_t\leftarrow
\operatorname{LowConfidence}(\mathbf{P}^{(1)},M_t,t)$
\STATE $\mathbf{P}^{(2)}\leftarrow
f_{\theta}\bigl(
\mathbf{C},
\operatorname{Remask}(\bar{\mathbf{y}},R_t)
\bigr)$
\STATE Update $\theta$ using
$\mathcal{L}(\mathbf{P}^{(1)},\mathbf{P}^{(2)},
\mathbf{y},M_t,V_t)$

\STATE \textbf{Inference}
\STATE $\hat L\leftarrow\operatorname{CLIPLen}(I^{1},I^{2})$;
$\hat{\mathbf{y}}^{(K)}
\leftarrow[\texttt{Mask}]^{\hat L}$
\FOR{$k=K,\ldots,1$}
    \STATE $\mathbf{P}^{(k)}
    \leftarrow f_{\theta}
    (\mathbf{C},\hat{\mathbf{y}}^{(k)})$
    \STATE $\bar{\mathbf{y}}\leftarrow
    \operatorname*{arg\,max}\mathbf{P}^{(k)}$
    \STATE $\hat{\mathbf{y}}^{(k-1)}
    \leftarrow\operatorname{Remask}
    \bigl(
    \bar{\mathbf{y}},
    \operatorname{LowConfidence}(\mathbf{P}^{(k)},k)
    \bigr)$
\ENDFOR
\FOR{$r=1,\ldots,R_{\mathrm{polish}}$}
    \STATE $\hat{\mathbf{y}}^{(0)}
    \leftarrow\operatorname{Polish}
    (\mathbf{C},\hat{\mathbf{y}}^{(0)})$
\ENDFOR
\RETURN $\operatorname{Decode}(\hat{\mathbf{y}}^{(0)})$
\end{algorithmic}
\end{algorithm}

\section{Experiments}
\label{sec:experiments}

\subsection{Experimental Setup}

\paragraph{Dataset and evaluation tasks.}
We evaluate EchoChange on the RSCC benchmark, which contains 62,351 bi-temporal remote-sensing image pairs paired with disaster-change descriptions. The average reference length is 72 tokens, making the benchmark substantially more demanding than short-form change captioning: a prediction must simultaneously preserve changed entities, damage events, quantities, directions, and spatial relations over a relatively long sequence. We use the official split for standard caption generation and evaluate 3,119 test records.

Beyond ordinary generation, we construct three controlled diagnostic tasks. The \emph{clean-control} set contains 892 correct descriptions and tests whether a model preserves already valid content. The correction benchmark contains 2,291 single-error samples and 553 multi-error samples; corruptions cover change type, direction, event, object, quantity, and relation. The recovery benchmark contains 892 samples in which factual spans are masked and must be reconstructed from the image pair and the remaining caption. These protocols separately test generation quality, active error correction, resistance to unnecessary rewriting, and fragment-level reconstruction.

All experiments were conducted on a system running Ubuntu 22.04, equipped with six NVIDIA A100 GPUs (40 GB each), using PyTorch 2.10.0 and CUDA 12.8.

\paragraph{Baselines and metrics.}
We compare with general-purpose multimodal models and remote-sensing-specific models. Standard generation is evaluated using ROUGE-L, METEOR, and Sentence-T5 semantic similarity (ST5-SCS). For correction, Clean Keep measures exact preservation of correct captions; Strict Hit requires every annotated error in a sample to be corrected; CCSR additionally requires non-target content to remain unchanged; Non-target Preservation (NTP) measures contextual stability; and error-level precision and F1 assess the accuracy of individual edits. Recovery is evaluated by masked-token accuracy (MTA), masked-span exact match (MSEM), full-caption token accuracy (FCTA), unresolved-mask rate (UMR), length mean absolute error (L-MAE), and repetition rates.

\subsection{Main Captioning Results}

\begin{table}[t]
\centering
\begingroup
\small
\setlength{\tabcolsep}{3.5pt}

\begin{tabular*}{\columnwidth}{
    @{\extracolsep{\fill}}lcccc@{}
}
\toprule
\multirow{2}{*}{\textbf{Model}} &
\multirow{2}{*}{\textbf{Size}} &
\multicolumn{2}{c}{\textbf{Lexical}} &
\multicolumn{1}{c}{\textbf{SCS}} \\
\cmidrule(lr){3-4}
\cmidrule(lr){5-5}
& &
R-L $\uparrow$ &
MET $\uparrow$ &
ST5 $\uparrow$ \\
\midrule

\multicolumn{5}{c}{\textbf{Open-source general models}} \\
\midrule
BLIP-3           & 3B  & 4.53  & 10.85 & 44.05 \\
Kimi-VL          & 3B  & 12.47 & 16.95 & 51.35 \\
Phi-4-Multimodal & 4B  & 4.09  & 1.45  & 34.55 \\
Qwen2-VL         & 7B  & 11.02 & 9.95  & 45.55 \\
LLaVA-NeXT       & 8B  & 12.51 & 13.29 & 46.99 \\
LLaVA-OneVision  & 8B  & 8.40  & 10.97 & 46.15 \\
InternVL3        & 8B  & 12.76 & 15.77 & 51.84 \\
Pixtral          & 12B & 12.34 & 15.94 & 49.36 \\
\midrule

\multicolumn{5}{c}{\textbf{Remote-sensing-specific models}} \\
\midrule
CCExpert & 7B & 7.61 & 4.32  & 40.81 \\
TEOChat  & 7B & 7.86 & 5.77  & 52.64 \\
RSCCM    & 7B & \underline{14.99} & 16.05 & 58.52 \\
\midrule

\multicolumn{5}{c}{\textbf{Base model and our method}} \\
\midrule
Qwen3.5 (Base)
& 9B
& 12.89
& \underline{19.04}
& \underline{64.09} \\

\addlinespace[3pt]

\textbf{EchoChange (Ours)}
& 9B
& \begin{tabular}[t]{@{}c@{}}
    \textbf{26.18} \\
    (+11.19)
  \end{tabular}
& \begin{tabular}[t]{@{}c@{}}
    \textbf{30.86} \\
    (+11.82)
  \end{tabular}
& \begin{tabular}[t]{@{}c@{}}
    \textbf{77.40} \\
    (+13.31)
  \end{tabular}
\\
\bottomrule
\end{tabular*}
\endgroup

\caption{Comparison with general-purpose and remote-sensing-specific
models. Qwen3.5-9B, the base model of EchoChange, is reported separately
for direct comparison. R-L, MET, and ST5 denote ROUGE-L, METEOR, and
ST5-SCS, respectively. The best results are shown in bold, while the
strongest baseline results are underlined. Parenthetical values indicate
the absolute improvements over the strongest baseline for each metric.}
\label{tab:performance_comparison}
\end{table}

EchoChange consistently outperforms both general-purpose MLLMs and remote-sensing-specialized baselines. Its absolute advantage ranges from 11.19 to 18.88 points across the three metrics, rather than being confined to either surface-level word overlap or semantic similarity. The particularly large ST5-SCS improvement indicates that the generated descriptions better preserve the overall disaster-change meaning, while the simultaneous ROUGE-L and METEOR gains show that this semantic advantage is accompanied by more accurate entities, actions, and relations. These results support the central premise of EchoChange: an editable, globally denoised answer is well suited to long-form, fact-intensive description of ordered bi-temporal evidence.

\subsection{Factual Correction and Preservation}

\begin{table*}[t]
  \centering
  \begingroup
  \setlength{\tabcolsep}{2pt}
  \small
  \begin{tabular*}{\textwidth}{
    @{\extracolsep{\fill}}lccccccccccc@{}
  }
    \toprule
    \multirow{2}{*}{\textbf{Model}} &
    \textbf{Clean} &
    \multicolumn{5}{c}{\textbf{Single Error}} &
    \multicolumn{5}{c}{\textbf{Multiple Errors}} \\
    \cmidrule(lr){2-2}
    \cmidrule(lr){3-7}
    \cmidrule(lr){8-12}
    & \textbf{Keep $\uparrow$}
    & \textbf{Strict $\uparrow$}
    & \textbf{CCSR $\uparrow$}
    & \textbf{NTP $\uparrow$}
    & \textbf{Err.-P $\uparrow$}
    & \textbf{Err.-F1 $\uparrow$}
    & \textbf{Strict $\uparrow$}
    & \textbf{CCSR $\uparrow$}
    & \textbf{NTP $\uparrow$}
    & \textbf{Err.-P $\uparrow$}
    & \textbf{Err.-F1 $\uparrow$} \\
    \midrule
    Qwen2-VL-7B
    & \textbf{88.00}
    & 5.11 & 2.97 & \textbf{67.53} & 5.06 & 5.12
    & 0.36 & 0.00 & \textbf{63.11} & 9.98 & 7.11 \\
    InternVL3-8B
    & 52.24
    & \underline{27.50} & \textbf{16.67} & 38.59
    & 12.84 & 17.57
    & \underline{7.23} & \underline{4.16} & 31.28
    & 21.11 & \underline{23.74} \\
    RSCCM
    & 36.55
    & 17.33 & 6.55 & 29.29 & 3.51 & 5.86
    & 4.70 & 1.99 & 20.98 & 7.10 & 10.49 \\
    Qwen3.5-9B
    & \underline{83.41} 
    & 18.68 & 14.01 & \underline{65.95} & \underline{20.03} & \underline{19.38}
    & 4.16 & 2.53 & \underline{60.40} & \underline{32.07} & 22.75\\
    \textbf{EchoChange (Ours)}
    & 79.26
    & \textbf{65.34} & \underline{16.15} & 23.26
    & \textbf{30.47} & \textbf{41.59}
    & \textbf{41.23} & \textbf{9.04} & 18.08
    & \textbf{42.67} & \textbf{51.26} \\
    \bottomrule
  \end{tabular*}
  \endgroup
  \caption{Clean-caption preservation and factual correction. Strict denotes sample-level all-target correction; CCSR requires semantic correction and non-target preservation; NTP measures preservation outside the injected positions; Err.-P/F1 are error-instance-level metrics. All values are percentages.}
  \label{tab:correction}
\end{table*}

The correction task exposes a distinction that is obscured by ordinary caption scores: a model must recognize an error, replace it with the visually supported fact, and avoid damaging valid context as shown in Table ~\ref{tab:correction}. EchoChange attains strict-hit rates of 65.34\% and 41.23\% on single- and multi-error descriptions, respectively, and more than doubles the strongest baseline error-level F1 in both settings. The drop from single-error to multi-error settings is expected because strict success requires all corrupted facts in a caption to be repaired simultaneously; nevertheless, EchoChange retains a much wider margin under this harder condition. This behavior is consistent with the purpose of dual-pass training: the model learns from a self-generated draft whose visible tokens are not guaranteed to be correct, rather than learning only to fill masked positions surrounded by ground-truth context.

The clean control reveals the complementary side of draft revision. Qwen2-VL preserves more untouched captions (88.00\%), but seldom performs a correct edit when an error is present. EchoChange retains 79.26\% of clean descriptions while remaining substantially more responsive to actual errors, showing that revision was not obtained by indiscriminately rewriting every input. This correction--preservation balance matches the asymmetric supervision used in training: erroneous model drafts receive ground-truth correction, whereas lower-weight visible-token supervision discourages changes to already valid context. The lower NTP and near-tied single-error CCSR also expose the remaining boundary: edit locality is not yet perfect, and confidence-aware selection can still introduce secondary changes.

EchoChange's advantage is strongest for visually grounded factual variables. In Table ~\ref{tab:error_types}, relation errors are corrected at over 95\%, and event, object, and quantity corrections remain robust when several corruptions coexist. This pattern indicates that the model is not merely applying common linguistic substitutions; it can repeatedly consult bi-temporal evidence to resolve what changed, where it changed, and how many objects were involved. Change type remains the main exception: InternVL3 achieves 34.40\% on the single-error split, compared with 21.87\% for EchoChange. This weakness suggests that fine-grained lexical distinctions among semantically adjacent change verbs remain difficult even when the affected entity is correctly localized.

\begin{table}[t]
  \centering
  \begingroup
  \setlength{\tabcolsep}{1pt}
  \small
  \begin{tabular*}{\columnwidth}{
    @{\extracolsep{\fill}}lcccccc@{}
  }
    \toprule
    \textbf{Model}
    & \textbf{Chg.} $\uparrow$
    & \textbf{Dir.} $\uparrow$
    & \textbf{Evt.} $\uparrow$
    & \textbf{Obj.} $\uparrow$
    & \textbf{Qty.} $\uparrow$
    & \textbf{Rel.} $\uparrow$ \\
    \midrule
    Qwen2-VL-7B
    & 2.92
    & 3.77
    & 16.76
    & 2.09
    & 0.00
    & 9.80 \\

    InternVL3-8B
    & \textbf{34.40}
    & 30.19
    & \underline{30.64}
    & \underline{20.71}
    & 5.36
    & 58.82 \\

    RSCCM
    & 16.62
    & 11.32
    & 21.68
    & 9.23
    & 5.36
    & \underline{67.16} \\

    Qwen3.5-9B
    & \underline{28.57}
    & \textbf{62.26}
    & 25.43
    & 11.25
    & \underline{7.14}
    & 29.41 \\

    \textbf{EchoChange (Ours)}
    & 21.87
    & \underline{49.06}
    & \textbf{81.21}
    & \textbf{68.35}
    & \textbf{69.64}
    & \textbf{95.59} \\
    \bottomrule
  \end{tabular*}
  \endgroup

  \caption{Strict correction hit rates by error type. Chg., Dir., Evt.,
  Obj., Qty., and Rel. denote change, direction, event, object, quantity,
  and relation errors, respectively. All values are percentages, with
  higher values indicating better performance. The best and second-best
  results in each column are highlighted in bold and underlined,
  respectively.}
  \label{tab:error_types}
\end{table}

\subsection{Analysis of Two-Stage Inference}

\begin{table}[t]
  \centering
  \begingroup
  \setlength{\tabcolsep}{1pt}
  \small
  \begin{tabular*}{\columnwidth}{
    @{\extracolsep{\fill}}lccccc@{}
  }
    \toprule
    \textbf{Configuration}
    & \textbf{D}
    & \textbf{P}
    & \textbf{ROUGE-L $\uparrow$}
    & \textbf{METEOR $\uparrow$}
    & \textbf{ST5-SCS $\uparrow$} \\
    \midrule
    Short two-stage
    & 4
    & 4
    & 24.45
    & 25.39
    & 73.98 \\
    Polish only
    & 0
    & 20
    & 22.75
    & 21.61
    & 71.92 \\
    Denoise only
    & 16
    & 0
    & 25.20
    & 28.30
    & 75.49 \\
    \textbf{Full two-stage}
    & 16
    & 4
    & \textbf{26.18}
    & \textbf{30.86}
    & \textbf{77.40} \\
    \bottomrule
  \end{tabular*}
  \endgroup
  \caption{Effect of denoising and polishing steps on caption quality.
  D and P denote the respective numbers of denoising and polishing
  steps.}
  \label{tab:inference_ablation}
\end{table}

Table~\ref{tab:inference_ablation} shows that denoising provides the principal quality gain, whereas polishing is most effective after a globally coherent draft has been formed. Denoising alone reaches an ST5-SCS of 75.49, and adding four polishing steps raises METEOR by 2.56 points while producing the best result across all three metrics. By contrast, even twenty polishing steps without denoising remain inferior, indicating that repeated local revision cannot compensate for the absence of an initial stage that establishes the event category, changed entities, and their temporal relations. The comparison also suggests that polishing is not intended to reconstruct the caption from an unstructured answer state. Instead, it operates on an already meaningful draft, correcting residual lexical errors, incomplete phrases, and inconsistencies among factual elements. This supports a clear division of labor: denoising constructs the global change structure through confidence-guided masked-token prediction, while polishing performs late-stage draft refinement. The latter is not an unrelated post-processing module; it reuses the full-answer revision behavior learned through dual-pass remasking, where Pass~2 is conditioned on a model-generated draft rather than a reference-derived context.
\begin{figure}[!htbp]
    \centering
    \includegraphics[width=\linewidth]{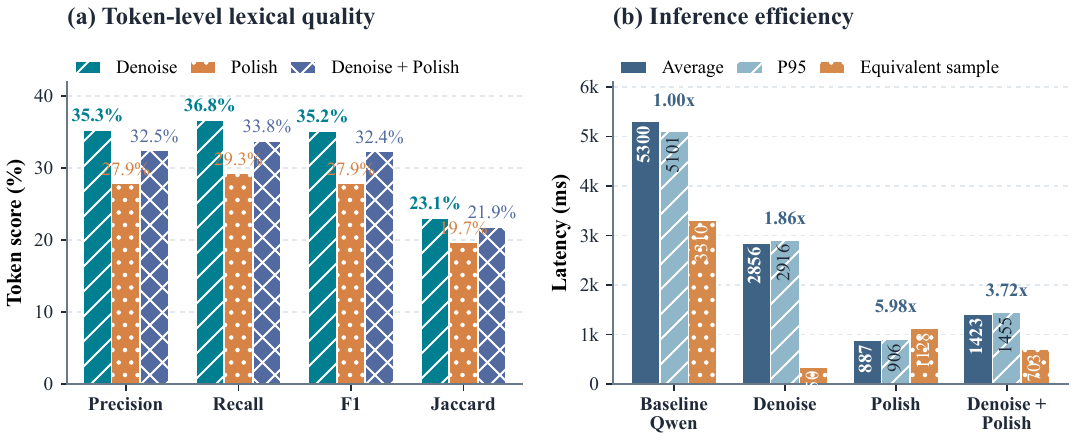}
    \caption{Token-level lexical quality and inference efficiency of the denoising and polishing regimes. (a) Precision, recall, F1, and Jaccard scores. (b) Average, P95, and equivalent-sample latency; labels above each group report the average speed-up relative to the baseline Qwen model.}
    \label{fig:inference_quality_efficiency}
\end{figure}

Figure~\ref{fig:inference_quality_efficiency} examines the two stages from lexical and computational perspectives. Denoising alone obtains the strongest token-level F1 (35.20\%) and Jaccard score (23.10\%); the combined procedure retains 32.40\% and 21.90\%, respectively, while reducing average latency from 5.30~s for the baseline Qwen model to 1.42~s, corresponding to a 3.72$\times$ speed-up. Polishing alone is faster but has lower lexical agreement, so speed cannot be considered independently of output quality. The combined setting offers the more useful operating point: it preserves most of the denoising-stage lexical evidence while realizing a substantial latency reduction. 

\subsection{Masked Factual-Span Recovery}

EchoChange recovers masked factual content much more reliably than models trained only for left-to-right continuation. Its masked-token accuracy reaches 72.32\%, a 35.21-point gain over the strongest baseline in Table ~\ref{tab:recovery}, and the corresponding span-level advantage shows that the improvement is not produced by recovering only easy isolated words. The high full-caption token accuracy and near-zero length error further indicate that newly reconstructed spans remain compatible with the visible context instead of changing the surrounding sentence structure. EchoChange also resolves every mask in the evaluation set. Its bigram and trigram repetition rates remain at 0.05\% and 0.00\%, respectively, showing that iterative refinement does not obtain its recovery gain by collapsing into repeated lexical patterns. Read together with correction and clean control, recovery separates the three behaviors learned by the answer-region objective: reconstruct missing evidence, revise incorrect visible evidence, and preserve correct evidence.

\begin{table}[t]
  \centering
  \begingroup
  \setlength{\tabcolsep}{0.5pt}
  \small
  \begin{tabular*}{\columnwidth}{
    @{\extracolsep{\fill}}lccccc@{}
  }
    \toprule
    \textbf{Model}
    & \textbf{MTA $\uparrow$}
    & \textbf{MSEM $\uparrow$}
    & \textbf{FCTA $\uparrow$}
    & \textbf{UMR $\downarrow$}
    & \textbf{L-MAE $\downarrow$} \\
    \midrule
    Qwen2-VL-7B
    & 18.87
    & 17.04
    & 74.02
    & 11.55
    & 8.14 \\
    InternVL3-8B
    & 22.32
    & 20.74
    & 72.79
    & \underline{0.11}
    & 13.58 \\
    RSCCM
    & \underline{37.11}
    & \underline{35.76}
    & 82.41
    & 3.25
    & \underline{1.03} \\
    Qwen3.5-9B
    & 12.04
    & 11.55
    & \underline{82.78}
    & 3.36
    & 1.90 \\
    \textbf{EchoChange (Ours)}
    & \textbf{72.32}
    & \textbf{68.83}
    & \textbf{95.10}
    & \textbf{0.00}
    & \textbf{0.07} \\
    \bottomrule
  \end{tabular*}
  \endgroup
  \caption{Masked-fragment recovery results. MTA, MSEM, FCTA, UMR,
  and L-MAE denote masked-token accuracy, masked-span exact match,
  full-caption token accuracy, unresolved-mask rate, and length MAE,
  respectively.}
  \label{tab:recovery}
\end{table}

\subsection{Qualitative Analysis of Progressive Generation and Model Comparison}
\begin{figure*}[!ht]
    \centering
    \includegraphics[width=\linewidth]{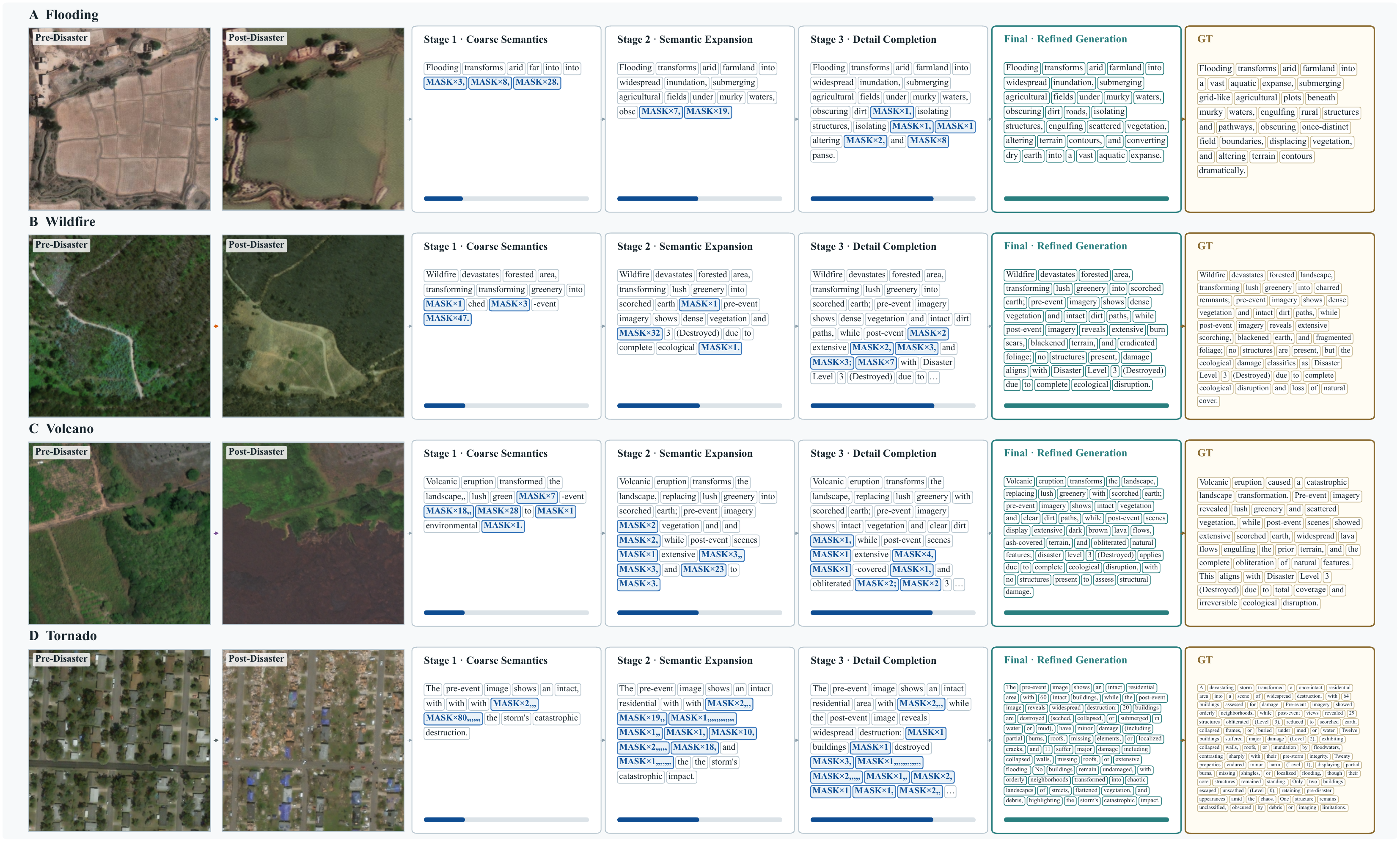}
    \caption{Visualization of the progressive generation process of EchoChange. For each bi-temporal image pair, the caption evolves from coarse event semantics to semantic expansion, detail completion, and final refinement. Masked positions are predicted jointly, while uncertain tokens remain editable across iterations, allowing unsupported intermediate hypotheses to be reconsidered using the global visual and textual context.}
    \label{fig:progressivegeneration}
\end{figure*}

\begin{figure*}[!ht]
    \centering
    \includegraphics[width=\linewidth]{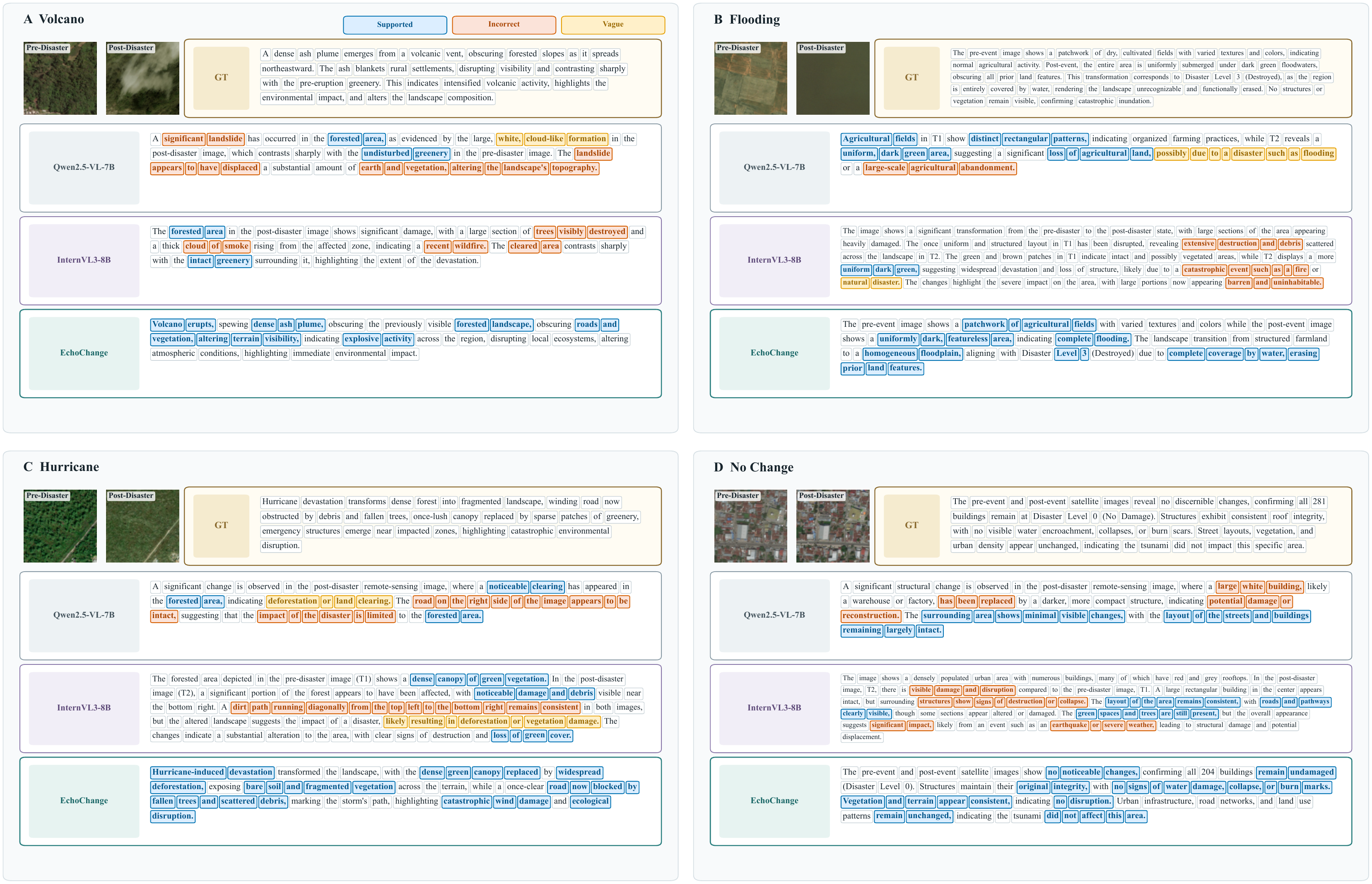}
    \caption{Qualitative comparison between EchoChange and autoregressive baselines under identical image pairs and instructions. Blue, orange, and yellow denote visually supported, incorrect, and vague spans, respectively. Compared with Qwen2-VL-7B and InternVL3-8B, EchoChange produces descriptions with more supported factual content and less unsupported elaboration, while the retained errors illustrate the remaining difficulty of ambiguous scenes and fine-grained quantity estimation.}
    \label{fig:compares}
\end{figure*}

Figure~\ref{fig:progressivegeneration} visualizes EchoChange's progressive generation: starting from a heavily masked sequence, the model establishes the dominant event and scene transition, then introduces affected objects, spatial relations, and damage details through successive iterations, with unsupported hypotheses remasked and revised as the global description develops. Figure~\ref{fig:compares} compares EchoChange with Qwen2-VL-7B and InternVL3-8B. Autoregressive baselines often extend inaccurate early interpretations with unsupported objects or avoid verifiable details through vague expressions; EchoChange instead produces more visually supported descriptions with less unsupported elaboration. Ambiguous visual evidence and fine-grained quantity estimation remain challenging. Together, these visualizations show how EchoChange's revisable generation improves factual grounding over conventional autoregressive decoding.

\section{Conclusion}

We presented EchoChange, a discrete diffusion framework for generating factual disaster change descriptions from bi-temporal remote-sensing images. Its dual-pass remasking objective equips the model with an explicit draft-revision capability: EchoChange learns not only to reconstruct masked factual content, but also to reassess visible tokens in its own imperfect drafts and preserve content that is already valid. During inference, confidence-guided denoising repeatedly redirects uncertain positions for reconsideration, while the subsequent polishing stage refines residual factual and lexical inconsistencies after the global description has formed. Experiments on RSCC show that EchoChange surpasses both general-purpose and remote-sensing-specific baselines in lexical and semantic caption quality. Factual-correction, clean-control, fragment-recovery, and denoise--polish evaluations further demonstrate its ability to repair single and multiple factual errors, reconstruct missing evidence, and avoid indiscriminate rewriting. 

\clearpage
\onecolumn
\bibliography{main}


\end{document}